\documentclass[runningheads]{llncs}
\usepackage{graphicx}
\usepackage{amsfonts}
\usepackage{amssymb}
\usepackage{siunitx}
\usepackage{float}
\usepackage{color}
\usepackage{array, makecell}
\usepackage{caption}
\usepackage{microtype}
\usepackage{float}
\usepackage{multirow,bigdelim,dcolumn,booktabs}
\usepackage{amsfonts}
\usepackage{makecell}
\usepackage{amsmath, bm}
\usepackage{color, colortbl}
\usepackage{hyperref}
\usepackage{url}
\usepackage{makecell}
\usepackage{graphicx}
\usepackage{multirow}
\usepackage{subcaption}
\usepackage{float}
\usepackage{caption}
\usepackage[export]{adjustbox} 

\begin{document}
\title{Enhancing Transformer Efficiency for Multivariate Time Series Classification}
\titlerunning{Time Series Classification}
%
\author{Yuqing Wang\inst{1} \and
Yun Zhao\inst{1} \and
Linda Petzold\inst{1}}

\renewcommand{\thefootnote}{\fnsymbol{footnote}}

\authorrunning{Y. Wang}
%
\institute{Department of Computer Science, University of California, Santa Barbara \\
\email{wang603@ucsb.edu}\\
}
\maketitle              
\begin{abstract}
Most current multivariate time series (MTS) classification algorithms focus on improving the predictive accuracy. However, for large-scale (either high-dimensional or long-sequential) time series (TS) datasets, there is an additional consideration: to design an efficient network architecture to reduce computational costs such as training time and memory footprint. In this work we propose a methodology based on module-wise pruning and Pareto analysis to investigate the relationship between model efficiency and accuracy, as well as its complexity. Comprehensive experiments on benchmark MTS datasets illustrate the effectiveness of our method.

\keywords{Deep learning \and Model efficiency \and Pareto analysis \and Time series classification.}
\end{abstract}

\section{Introduction}

Time series (TS) data is ubiquitous, occurring in healthcare~\cite{li2014physiological}, stock market~\cite{liu2020improved}, astronomy~\cite{fu2011review}, and many other domains ~\cite{gao2017physics,hu2020lightweight}. With the advance of sensing techniques, TS classification across wide-ranging domains has gained much interest during the past decade~\cite{fawaz2019deep,ruiz2021great}. 

The availability of the UCR/UEA time series benchmark datasets~\cite{ruiz2021great} has led to an abundance of TS classification algorithms~\cite{husken2003recurrent,zhao2017convolutional,lines2018time,dempster2020rocket,zerveas2021transformer}. The classification accuracy has been the key metric used to evaluate existing methods 
~\cite{lines2015time}. However, the high accuracy of these algorithms often comes with the cost of high computational complexity~\cite{schafer2016scalable}. From common preconceptions in natural language processing (NLP) and computer vision (CV), in order to achieve high accuracy, training top performing models with millions/billions of parameters is a computationally intensive task, requiring days or weeks on many parallel GPUs or TPUs. However, such intensive training makes the model difficult to retrain for further improvement on performance. Likewise, for large-scale time series data with high dimensionality or long sequence length, it is challenging to maintain the balance between the predictive accuracy and training efficiency.

In this work, we propose a method to investigate the relationship between model efficiency and its effectiveness, as well as its complexity for MTS classification. The model architecture is based on Transformer and Fourier transform. We use 18 benchmark MTS datasets for evaluation. Comprehensive experiments are conducted on all datasets, including ablation study of each module of the network and module-by-module pruning in terms of accuracy, training speed, and model size. Experimental results demonstrate the competitive performance of our proposed architecture compared with current state-of-the-art methods. Ablation studies identify the main contributors to the predictive performance, such as multi-head self-attention and Fourier transform. In addition, module-wise pruning of the network reveals the trade-off between model efficiency and effectiveness, as well as model efficiency and complexity. Finally, we conduct Pareto analysis to examine the trade-off between efficiency and performance.

The main contributions of this paper are highlighted as follows:

\begin{enumerate}
    \item[(1)] To the best of our knowledge, this is the first paper to perform Pareto analysis to investigate the relationship between efficiency and accuracy.
    \item[(2)] Through module-by-module pruning, comprehensive experimental results indicate an evident trade-off between model efficiency and its effectiveness, as well as its complexity.
    \item[(3)] We employ Pareto analysis to investigate the relationship between model efficiency and performance. Such analysis methods can provide general guidance for researchers on how to select efficient model configurations, which can be applied to any model architecture.
\end{enumerate}

The remainder of this paper is organized as follows. Section~\ref{ref} describes related work of Transformer and Fourier transform on time series analysis and existing methods on model efficiency improvement. The network architecture is outlined in Section~\ref{method}. Section~\ref{experiments} discusses datasets and experiments on 18 benchmark datasets, including ablation studies, module-wise pruning and Pareto efficiency visualization. Finally, our conclusions are presented in Section~\ref{con}.

\section{RELATED WORK}
\label{ref}

\textbf{Neural Networks for Time Series Classification.} Currently, most TS classification algorithms can be divided into three categories: feature-based~\cite{fulcher2014highly}, distance-based~\cite{abanda2019review}, and neural network based methods~\cite{fawaz2019deep}. Here, we focus only on neural network based methods. Since the advancements of deep learning, two popular frameworks, CNN and RNN, are widely applied in TS classification tasks.  ~\cite{wang2017time} combined Fully Convolutional Networks (FCN) and Residual Networks (ResNet) for univariate time series classification. \cite{lin2017gcrnn} developed a group-constrained method, which combines a CNN with an RNN. More recent works such as InceptionTime~\cite{fawaz2020inceptiontime}, TapNet~\cite{zhang2020tapnet}, and TST~\cite{zerveas2021transformer} are proposed for TS classification. For additional deep learning methods, we refer readers to ~\cite{fawaz2019deep}. \\

\noindent
\textbf{Fourier Transform in Time Series.}
The Fourier transform (FT) has been an important tool in time series analysis for decades~\cite{bloomfield2004fourier}, and is widely used for applications such as anomaly detection~\cite{ren2019time}, periodicity detection~\cite{puech2019fully}, and similarity measures~\cite{janacek2005likelihood}. The FT converts a TS from time domain to frequency domain, and uses Fourier coefficients to represent the original data. For the TS classification task, FTs have been used indirectly in disparate applications. For instance, ~\cite{geerken2005classifying} utilizes the FT to filter noisy data for vegetation type classification, and  ~\cite{samiee2014epileptic} uses the FT as a feature extraction technique to classify electroencephalography (EEG) data. However, none of the above methods apply the FT directly to TS classification, particularly in the context of neural networks. In contrast, we aim to apply the discrete FT and its inverse as modules of a deep learning framework. The unparameterized FT can reduce the computational cost of the network to some extent. \\

\noindent
\textbf{Transformer Networks for Time Series Classification.} With the exemplary performance of the Transformer architecture~\cite{vaswani2017attention} in NLP and CV, researchers in the time series community began exploring Transformers in TS classification in specific domains~\cite{oh2018learning,zhao2021bertsurv}. More recent works have generalized Transformer frameworks for MTS classification. ~\cite{zerveas2021transformer} adopts a Transformer encoder architecture for unsupervised representation learning of MTS. ~\cite{liu2021gated} explored an extension of the current Transformer architecture by gating, which merges two towers for MTS classification.  In contrast, we propose to generalize a mixing framework which utilizes both Transformer and FT. By replacing some self-attention sublayers with FT, the computational complexity can be reduced. \\

\noindent
\textbf{Model Training Efficiency.} Due to the increasing size of both models
and training data, many works have focused on improving model training efficiency through parameter reduction, such as DenseNet~\cite{huang2017densely} and EfficientNet~\cite{tan2019efficientnet}, training speed improvement including NFNets~\cite{brock2021high} and BotNet~\cite{srinivas2021bottleneck}, or both~\cite{tan2021efficientnetv2}. One of the most common techniques to improve network efficiency is model pruning. Early works focused on non-structured methods. For instance, \cite{lecun1990optimal,han2015learning} proposed to remove individual weight values. Recent works focused more on structured methods, such as channel weight pruning based on $l_1$ norm~\cite{li2016pruning}.

\section{METHODOLOGY}
\label{method}
\begin{figure*}[ht]
\centering
\includegraphics[width=1.05\linewidth]{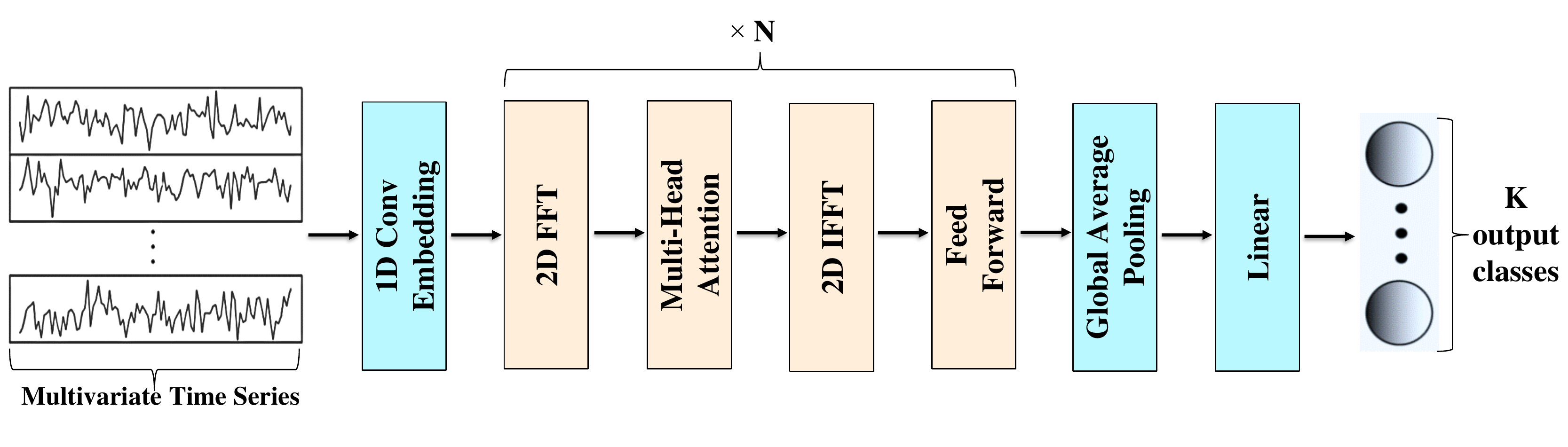}
\caption{An overview of the full model framework. Our architecture is based on Transformer and Fourier transform. Following the sequence embedding, we apply a 2D discrete Fourier transform (particularly Fast Fourier transform) to convert the TS features from the time domain to the frequency domain, a multi-head self-attention layer, and a 2D inverse discrete Fourier transform to map the features back to the time domain. Then we employ a Global Average Pooling (GAP) layer to average the output of the
MTS over the entire time dimension. Finally, a Softmax layer is used for the multi-class MTS classification task.}
\label{model}
\end{figure*}

In this section, we present our network architecture, which contains all of the modules for potential model pruning. The overall model structure is illustrated in Figure~\ref{model}.

\textbf{Input Embeddings.} Input embeddings are commonly used in NLP models, which map relatively low-dimensional vectors to  high-dimensional vectors to facilitate sequence modeling~\cite{kim-2014-convolutional}. Correspondingly, an embedding for TS sequence is required to capture the dependencies among different features without considering the temporal information~\cite{song2018attend}. Our framework employs a 1D convolutional layer to obtain the K-dimensional embeddings at each time step.

\textbf{Discrete Fourier Transform.}
The Fourier transform decomposes a function of time into its constituent frequencies. For clarity, we first consider the 1D Discrete Fourier transform (DFT). Given a sequence of complex numbers $x(n)$ with $0 \leq n \leq N -1$, the 1D DFT is defined by
\begin{align*}
    X(k) = \sum_{n=0}^{N-1} x(n) \cdot e^{-\frac{2\pi i}{N} kn} =  \sum_{n=0}^{N-1} x(n) \cdot W_N^{kn}, ~ 0 \leq k \leq N-1,
\end{align*}
where $W_N^{kn} = e^{-\frac{2\pi i}{N}kn}$.
Given the DFT $X(k)$, the original sequence can be recovered by the inverse DFT (IDFT)
\begin{align*}
    x(n) = \frac{1}{N} \sum_{k=0}^{N-1} X(k) \cdot e^{\frac{2\pi i}{N} kn}, ~ 0 \leq n \leq N-1.
\end{align*}

The 2D DFT is a direct extension of the 1D DFT, obtained by alternately performing the 1D DFT on the row and column dimensions. Given a 2D signal $x(m,n)$ with $0 \leq m \leq M -1, 0 \leq n \leq N -1$, the 2D DFT is given by
\begin{align*}
    X(k,l) = \sum_{m=0}^{M-1} \sum_{n=0}^{N-1} x(m,n) \cdot e^{-2\pi j (\frac{km}{M} + \frac{ln}{N})}.
\end{align*}
Similar to the 1D IDFT, the 2D DFT is invertible via the 2D IDFT,
\begin{align*}
    x(m,n) = \frac{1}{MN} \sum_{k=0}^{M-1} \sum_{l=0}^{N-1} X(k,l) \cdot e^{2\pi j (\frac{km}{M} + \frac{ln}{N})}.
\end{align*}

To compute the DFT efficiently, the Fast Fourier Transform (FFT) algorithm takes advantage of the periodicity and symmetry properties of $W_N^{kn}$ such that the computational complexity of the DFT reduces from $O(N^2)$ to $O(NlogN)$, regardless of dimension.

\textbf{Multi-head Attention.}
The multi-head attention (MHA) mechanism, the major component of the Transformer architecture~\cite{vaswani2017attention}, allows the model to jointly attend to information from different representation subspaces at different positions. MHA is defined as:
\begin{align*}
    {\rm MultiHead}(Q,K,V) = {\rm Concat}({\rm head}_1, {\rm head}_2, \cdots, {\rm head}_h) W^O,
\end{align*}
where $Q,K,V \in \mathbb{R}^{n \times d_{model}}$ are input embedding matrices, $n$ is the sequence length, $d_{model}$ is the embedding dimension, and $h$ is the number of heads. Each head $i$ is defined as:
\begin{align*}
    {\rm head}_i = {\rm Attention}(QW_i^Q,KW_i^K,VW_i^V) = {\rm softmax} \bigg( \frac{QW_i^Q(KW_i^K)^T}{\sqrt{d_k}} \bigg) VW_i^V,
\end{align*}
where $W_i^Q \in \mathbb{R}^{d_{model} \times d_k}, W_i^K \in \mathbb{R}^{d_{model} \times d_k}, W_i^V \in \mathbb{R}^{d_{model} \times d_v}, W_i^O \in \mathbb{R}^{hdv \times d_{model}}$ are parameter matrices to be learned.

\textbf{Global Average Pooling.} Global average pooling involves calculating the average value of all of the elements in a feature map. It is mainly used to reduce the amount of learnable parameters. 

\textbf{Batch Normalization.} Instead of using layer normalization in Transformer-related architectures in NLP,  we consider the necessity of applying batch normalization to each block shown in Figure~\ref{model}. Compared to layer normalization, batch normalization can mitigate the effect of outlier values in time series data, which does not appear in text representations.

\textbf{Activation Function.} Using the same activation function as the original Transformer architecture~\cite{vaswani2017attention}, we consider the necessity of applying the activation function \textit{gelu} for each module shown in Figure~\ref{model}.

\textbf{Feedforward Neural Network.} A position-wise feedforward neural network (FNN) is applied with two 1D convolutional layers with kernel size $1$, and a \textit{gelu} activation function in between.
    
\section{EXPERIMENTS}
\label{experiments}
In this section, we describe benchmark MTS datasets~\cite{ruiz2021great} used for experimental evaluation, the experimental setup, and corresponding results.

\subsection{DATASETS}
We select a set of 18 publicly available benchmark datasets
from the UCR/UEA classification archive: AtrialFibrillation (AF), BasicMotions (BM), Cricket (CR), DuckDuckGeese (DDG), Epilepsy (EP), EthanolConcentration (EC), ERing (ER), FingerMovements (FM), HandMovementDirection (HMD), Handwriting (HW), Heartbeat (HB), Libras (LIB), NATOPS (NATO), PEMS-SF (PEMS), RacketSports (RS), SelfRegulationSCP1 (SRS1), SelfRegulationSCP2 (SRS2), and UWaveGestureLibrary (UW). The main characteristics of each dataset are summarised in Table~\ref{datasummery}. All of the datasets have been split into training and testing sets by default. Thus, there are no preprocessing steps for these data. The predictive performance on all datasets is evaluated in terms of accuracy.

\begin{table}[ht]
\centering
\caption{Summary of the 18 UCR/UEA datasets used in experimentation.}
\scalebox{0.8}{
\begin{tabular}{c c c c c c c}
\Xhline{1.2pt}
\textbf{Dataset}  & \textbf{Code} & \textbf{Train Size} & \textbf{Test Size} & \textbf{Dimensions} & \textbf{Length} & \textbf{Classes}  \\ 
\hline
AtrialFibrillation & AF & 15 & 15 & 2 & 640 & 3 \\ 
BasicMotions & BM & 40 & 40 & 6 & 100 & 4 \\
Cricket & CR & 108 & 72 & 6 & 1197 & 12 \\
DuckDuckGeese & DDG & 50 & 50 & 1345 & 270 & 5 \\ 
Epilepsy & EP & 137 & 138 & 3 & 206 & 4 \\
EthanolConcentration & EC & 261 & 263 & 3 & 1751 & 4 \\
ERing & ER & 30 & 270 & 4 & 65 & 6 \\
FingerMovements & FM & 316 & 100 & 28 & 50 & 2 \\
HandMovementDirection & HMD & 160 & 74 & 10 & 400 & 4 \\
Handwriting & HW & 150 & 850 & 3 & 152 & 26 \\
Heartbeat & HB & 204 & 205 & 61 & 405 & 2 \\
Libras & LIB & 180 & 180 & 2 & 45 & 15 \\
NATOPS & NATO & 180 & 180 & 24 & 51 & 6 \\
PEMS-SF & PEMS & 267 & 173 & 963 & 144 & 7 \\
RacketSports & RS & 151 & 152 & 6 & 30 & 4 \\
SelfRegulationSCP1 & SRS1 & 268 & 293 & 6 & 896 & 2 \\
SelfRegulationSCP2 & SRS2 & 200 & 180 & 7 & 1152 & 2 \\
UWaveGestureLibrary & UW & 120 & 320 & 3 & 315 & 8 \\

\Xhline{1.2pt}
\end{tabular}}
\label{datasummery}
\end{table}

\vspace{-1cm}
\subsection{SETUP}
We set aside $20\%$ of the default training set for the validation set, which we used to select the best collection of hyperparameters.
All experiments were implemented in Pytorch~\cite{paszke2019pytorch} on one GTX 1080 Ti GPU. We minimized the cross entropy loss with the Adam~\cite{kingma2014adam} optimizer for training. The hyperparameter search space for each dataset is listed in Table~\ref{hyperparameters}. Note that the batch size choice is limited by the available GPU memory.

\begin{table}[ht] 
\centering
\caption{Hyperparameter search space of the model on each dataset. If the number of layers of a module is equal to 0, then this module is removed in the pruned model.}
\scalebox{0.9}{
\begin{tabular}{c c} 
\Xhline{1.5pt}
Hyperparameters     & Search Space  \\ 
\hline
learning rate       & [1e-3, 5e-3, 1e-4, 5e-4, 1e-5, 5e-5]    \\
dropout rate        & [0.1, 0.2, 0.3]      \\
batch size          & [8, 16, 32]   \\
$\sharp$ of heads  & [4, 8, 16] \\
$\sharp$ of FFT layers & [0, 1, 2, 3, 4] \\
$\sharp$ of IFFT layers & [0, 1, 2, 3, 4] \\
$\sharp$ of MHA layers & [0, 1, 2, 3, 4] \\
$\sharp$ of Feedforward layers & [0, 1, 2, 3, 4] \\
\Xhline{1.5pt}
\end{tabular}}
\label{hyperparameters}
\end{table}

\begin{table}[H]
\centering
\caption{Ablation study in the testing accuracy loss on 18 datasets by removing each module at a time while leaving others the same. Each experiment is conducted $5$ times with different random seeds. The results are shown in the format of mean and standard deviation. Column $2$ shows the accuracy of the full model with all modules included. Columns $3$ to $10$ represent the accuracy when the module in that column is removed from the model. Bold indicates that the module contributes most to the loss in accuracy and underlining indicates that the module contributes least to the loss in accuracy when the module is removed.}
\begin{tabular}{c |c c c c c c c c c c}
\Xhline{1.2pt}
\textbf{Dataset}  & \textbf{Acc.} & \textbf{Unpruned} & \textbf{EMBED} & \textbf{FFT} & \textbf{IFFT} & \textbf{MHA} & \textbf{FFN} & \textbf{GAP} & \textbf{BN} & \textbf{ACT}  \\ 
\hline
\multirow{2}{*}{AF} & Mean & 0.667 & 0.600 & \textbf{0.400}  & 0.467 & \textbf{0.400} & \underline{0.667} & 0.533 & 0.600 & \underline{0.667} \\
& Std. & 0.003 & 0.005 &  0.005 &  0.004 &  0.003 & 0.006 &  0.006 &  0.004 & 0.003\\ 
\hline
\multirow{2}{*}{BM} & Mean & 0.975 & \underline{0.950}  & \textbf{0.725}  & 0.775   & 0.750 & 0.900 & 0.925  & 0.900  & \underline{0.950}  \\
& Std. & 0.008  & 0.010 & 0.012 & 0.009  & 0.012 & 0.010 &  0.014 & 0.009 & 0.011 \\
\hline
\multirow{2}{*}{CR} & Mean & 0.987 & 0.958  & 0.875 & 0.861  & \textbf{0.833} & 0.889  & 0.944 & \underline{0.972} & 0.944 \\
& Std. & 0.007 & 0.009 & 0.012 &  0.008 & 0.012 &  0.006 &  0.009 & 0.012 & 0.008 \\
\hline
\multirow{2}{*}{DDG} & Mean & 0.580  & \underline{0.580} & 0.440  & 0.420  & \textbf{0.380}  & 0.520 & 0.560 & 0.560  & \underline{0.580}  \\ 
& Std. & 0.016 &  0.017 & 0.020 &  0.016 & 0.014 &  0.016 & 0.016 & 0.014 &  0.016 \\ 
\hline
\multirow{2}{*}{EP} & Mean & 0.986  & \underline{0.978}  & \textbf{0.891} & 0.913 & 0.899  & 0.949 & 0.971 & 0.956 & 0.971  \\
& Std. & 0.014 & 0.013 & 0.016 &  0.014 & 0.014 & 0.012 & 0.014 &  0.013 & 0.015 \\
\hline
\multirow{2}{*}{EC} & Mean & 0.456 & \underline{0.445} & 0.376 & 0.395  & \textbf{0.365}  & 0.418 & 0.441 & \underline{0.445} & 0.452 \\
& Std. & 0.003  & 0.002 & 0.003 &  0.003 & 0.004 & 0.002 &  0.004 & 0.003 & 0.002 \\
\hline
\multirow{2}{*}{ER} & Mean & 0.963  & \underline{0.956}  & 0.896 & 0.889   & \textbf{0.885}  & 0.892  & 0.948 & 0.952  & \underline{0.956} \\
& Std. & 0.006 &  0.007 & 0.006 &  0.006 &  0.008 &  0.005 &  0.006 & 0.007 &  0.005 \\
\hline
\multirow{2}{*}{FM} & Mean & 0.640 & \underline{0.620} & \textbf{0.490}   & 0.520  & 0.500 & 0.600  & 0.590   & 0.610  & \underline{0.620} \\
& Std. & 0.009 & 0.008 &  0.007 &  0.008 & 0.010 & 0.008 &  0.009 & 0.010 &  0.011 \\
\hline
\multirow{2}{*}{HMD} & Mean & 0.486 & 0.446 & 0.365 & 0.351  & \textbf{0.338} & 0.406 & 0.459  & 0.432 & \underline{0.473} \\
& Std. & 0.018 & 0.016 & 0.020 &  0.017 & 0.018 & 0.019 & 0.018 & 0.016 & 0.020 \\
\hline
\multirow{2}{*}{HW} & Mean & 0.529  & \underline{0.514}  & 0.471 & 0.473  & \textbf{0.468} & 0.506 & 0.506 & 0.512  & \underline{0.514} \\
& Std. &  0.006 & 0.007 & 0.006 &  0.005 & 0.007 & 0.007 &  0.008 & 0.007 & 0.006\\
\hline

\multirow{2}{*}{HB} & Mean & 0.771 & \underline{0.766} & \textbf{0.683} & 0.707 & 0.688 & 0.751 & 0.756  & \underline{0.766} & 0.756 \\
 & Std. & 0.014 & 0.015 & 0.014 &  0.017 &  0.015 & 0.016 &  0.014 & 0.015 & 0.016 \\
\hline
\multirow{2}{*}{LIB} & Mean & 0.917  & 0.906  & \textbf{0.822} & 0.827 & 0.839 & 0.889  & 0.894 & 0.906 & \underline{0.911} \\
& Std. & 0.009 &  0.011 & 0.012 & 0.010 & 0.012 &  0.013 & 0.011 & 0.009 & 0.010\\
\hline
\multirow{2}{*}{NATO} & Mean & 0.844  & \underline{0.833} & \textbf{0.728}  & 0.739  & 0.750  & 0.772  & 0.811 & \underline{0.833} & \underline{0.833} \\
& Std. & 0.005 & 0.004 & 0.005 &  0.007 & 0.006 & 0.005 & 0.006 & 0.004 & 0.006 \\
\hline
\multirow{2}{*}{PEMS} & Mean & 0.908 & 0.884 & 0.815 & 0.809 & \textbf{0.803} & 0.867 & 0.879 & \underline{0.896} & \underline{0.896} \\
& Std. & 0.013 & 0.012 & 0.014 &  0.016 & 0.014 &  0.013 & 0.013 & 0.014 & 0.012\\
\hline
\multirow{2}{*}{RS} & Mean & 0.914 & 0.901 & \textbf{0.796}  & 0.816 & 0.803 & 0.855  & \underline{0.908} & 0.901  & \underline{0.908} \\
& Std. & 0.021 & 0.020 & 0.020 &  0.018 & 0.019 & 0.021 & 0.020 & 0.021 & 0.019 \\
\hline
\multirow{2}{*}{SRS1} & Mean & 0.915 & 0.894  & 0.836 & 0.823  & \textbf{0.819} & 0.853   & 0.887  & 0.894   & \underline{0.901}   \\
& Std. & 0.005 &  0.007 & 0.006 &  0.006 & 0.005 &  0.007 & 0.006 & 0.005 & 0.005 \\
\hline
\multirow{2}{*}{SRS2} & Mean & 0.600 & \underline{0.594} & \textbf{0.522} & 0.533   & 0.516  & 0.578 & 0.583  & 0.588 & \underline{0.594} \\
& Std. & 0.002 & 0.003 & 0.002 & 0.001 & 0.004 &  0.002 & 0.003 & 0.003 & 0.002 \\
\hline
\multirow{2}{*}{UW} & Mean & 0.922  & \underline{0.906} & 0.844  & 0.850  & \textbf{0.841}  & 0.875  & 0.894  & 0.897  & 0.903\\
& Std. & 0.006 & 0.008 & 0.009 & 0.006 & 0.007 &  0.008 & 0.006 & 0.007 & 0.007 \\

\Xhline{1.2pt}
\end{tabular}
\label{ablation}
\end{table}

\subsection{MODULE SETTINGS}
Based on Section~\ref{method}, we define the following eight modules of the network for further analysis: input embedding (EMBED), fast Fourier transform (FFT), inverse fast Fourier transform (IFFT), multi-head attention (MHA), feedforward neural network (FFN), global average pooling (GAP), batch normalization (BN), and activation function (ACT). The corresponding abbreviations of each module are shown in parentheses.

\subsection{ABLATION STUDY}
\label{as}
First, we conduct ablation studies to analyze the contributions of each module on the predictive performance. The contribution of each module is obtained when a module is removed from the full network while other modules remain intact. The fine-tuned results on 18 datasets are shown in Table~\ref{ablation}. Starting from Column $4$, the smaller the accuracy is, the larger the module's contribution is, and vice versa. The accuracy of each dataset for the unpruned model (Table~\ref{ablation} Column $3$) is competitive with current state-of-the-art methods~\cite{ruiz2021great}. Among eight modules, it can be seen that MHA and FFT contribute most to the predictive performance on $10$ out of the $18$ datasets and $9$ out of the $18$ datasets, respectively. For MTS data, the correlations between different dimensions across all time steps are important to consider. Hence, the MHA is able to catch different feature correlations, and influence the accuracy to a large extent. The FFT, as the core of signal processing and more generalized time series, extracts frequency information embedded in data, which provides a more straightforward representation compared to the original data in the time-domain. In contrast, we observe that EMBED, BN, and ACT contribute least to the predictive performance on $11$ out of the $18$ datasets, $5$ out of the $18$ datasets, and $13$ out of the $18$ datasets, respectively. Although these operations are important for the training of the model, they influence the testing accuracy marginally compared with MHA and FFT.

\begin{figure}[htbp]
    \centering
    \subfloat[]{\includegraphics[width=0.48\textwidth,]{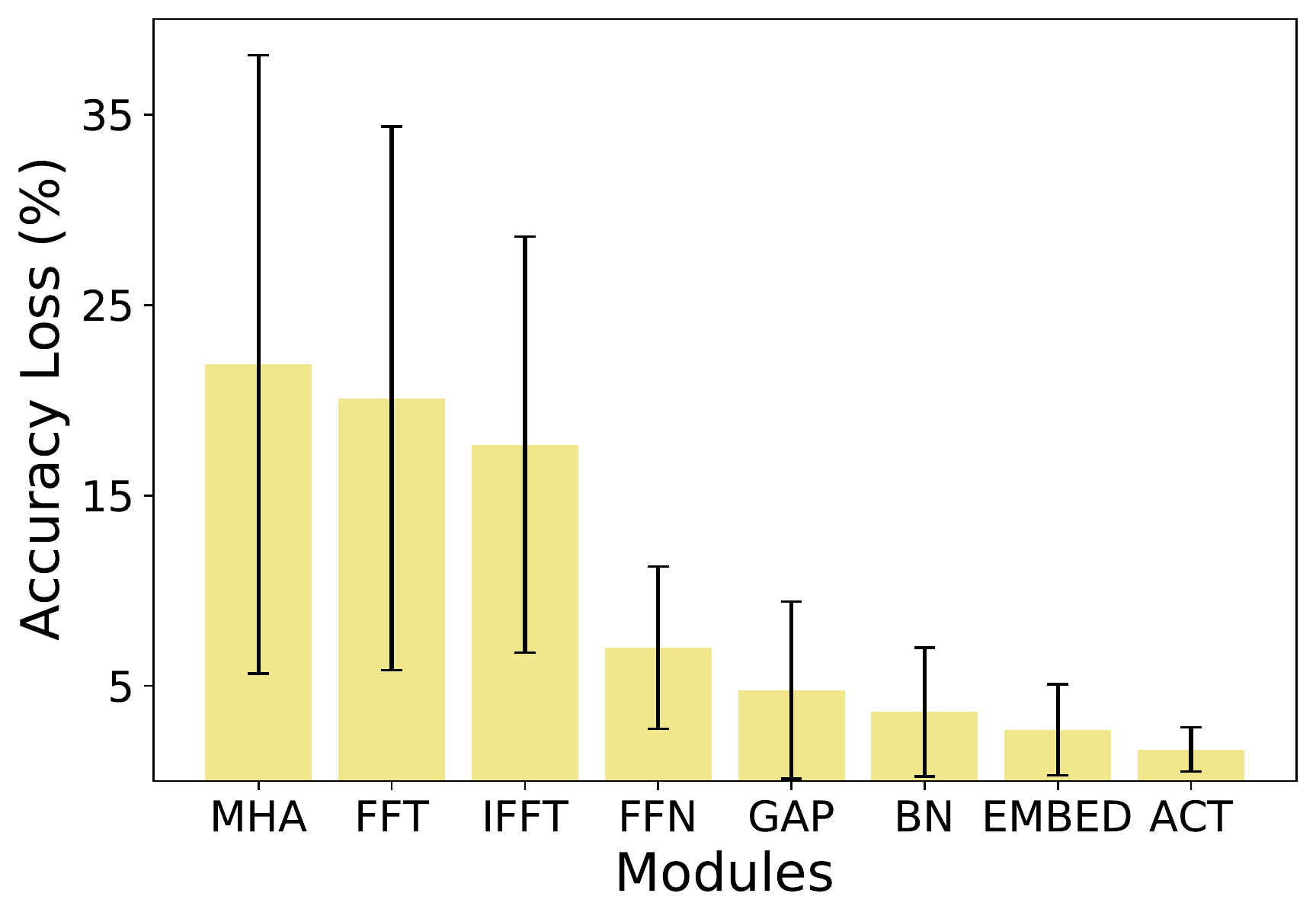}\label{acc_lossa}}
    \hfill
    \subfloat[]{\includegraphics[width=0.48\textwidth, ]{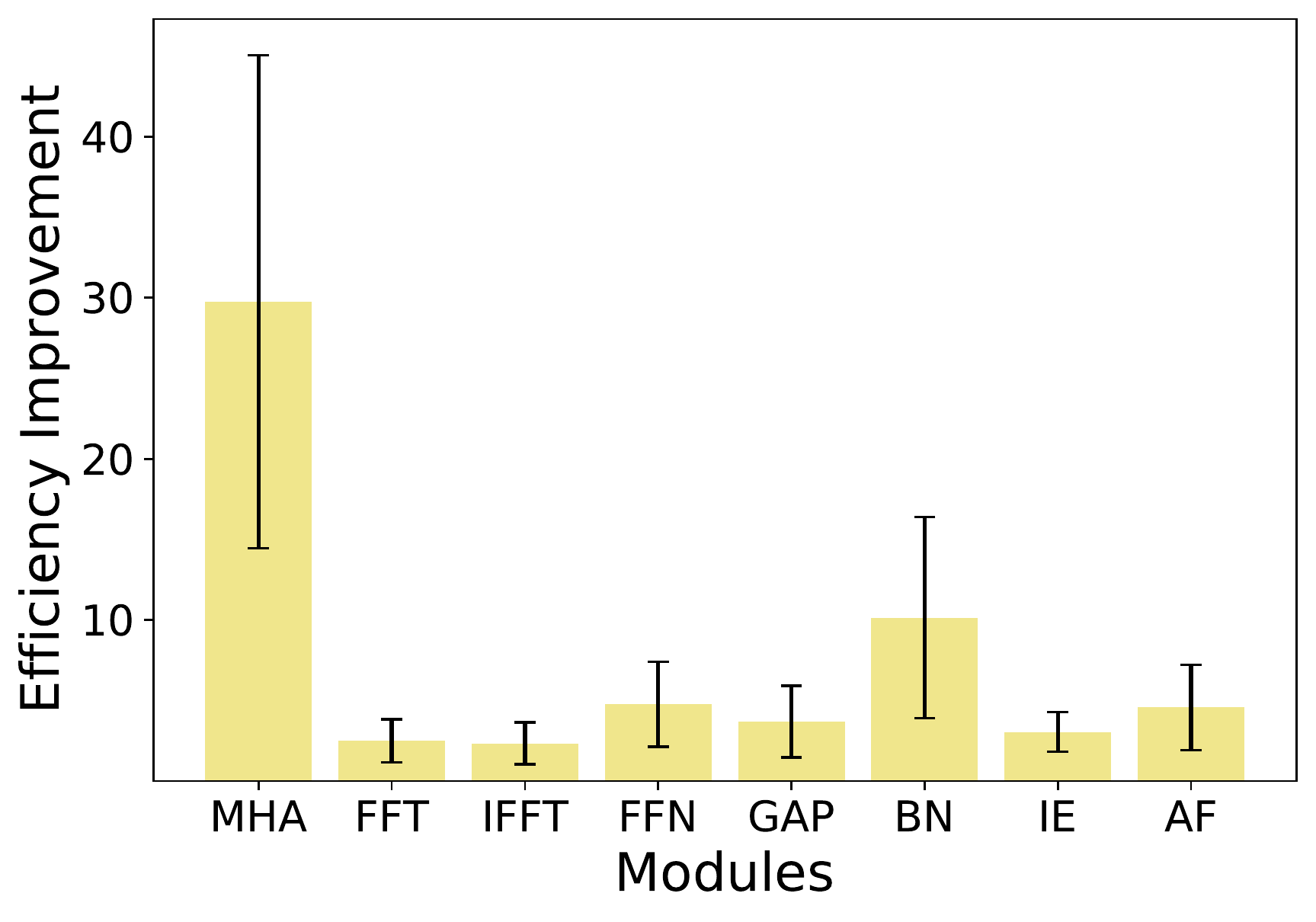}\label{acc_lossb}}
  \caption{(a) represents the average testing accuracy loss across all datasets while removing one module at a time and other modules remain in the network. Modules MHA, FFT, and IFFT bring about larger influence on the predictive performance due to the high percentage of accuracy loss when removing them. In comparison, BN, EMBED, and ACT bring about marginal influence on the predictive performance compared with other modules. (b) represents the corresponding average efficiency improvement across all datasets when one module is removed from the network while other modules keep intact.}
  \label{acc_loss}
\end{figure}

To clearly demonstrate the influence of each module on the predictive performance and efficiency of the network, the averaged testing accuracy loss and the corresponding efficiency improvement for each module (compared with the unpruned model) over all datasets are presented in Figure~\ref{acc_loss}. Here, efficiency is defined as the product of training time per epoch and the amount of learnable parameters. The higher the product, the lower the efficiency is. In consideration of highly diversed datasets with respect to sequence length, number of samples, and dimensionality, the average loss in accuracy for each module demonstrates a high variance from Figure~\ref{acc_lossa} as the performance loss extent can vary depending on dataset characteristics. The modules MHA, FFT, and IFFT demonstrate a notable influence on the model performance on average (21.9\%, 20.1\%, and 17.7\% loss in accuracy respectively). For modules like BN, EMBED and ACT, removing them bring about minimal accuracy loss compared to other modules (3.6\%, 2.7\%, and 1.6\% respectively). Meanwhile, comparing Figure~\ref{acc_lossa} and Figure~\ref{acc_lossb}, the module which has larger impact on the predictive performance does not indicate that removing it can bring about more efficiency improvement. For instance, the computationally inexpensive FFT influences the predictive performance to a large extent. In contrast, although the computational cost of BN is high, its contribution to the performance is marginal.

\subsection{MODULE-BY-MODULE PRUNING}
\begin{figure}[htbp]
\centering
 \subfloat[]{\includegraphics[width=0.5\textwidth,]{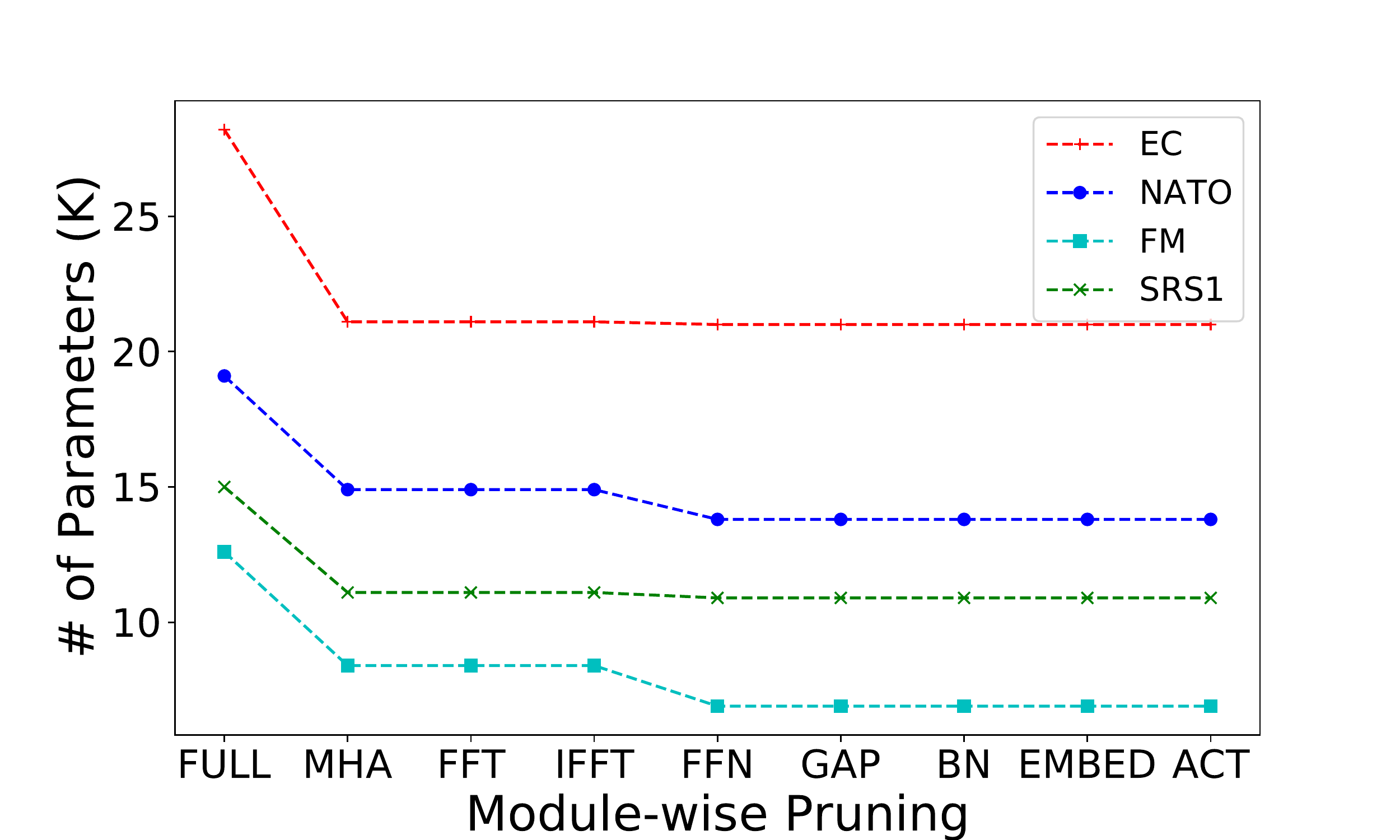}}
    \hfill
    \subfloat[]{\includegraphics[width=0.5\textwidth, ]{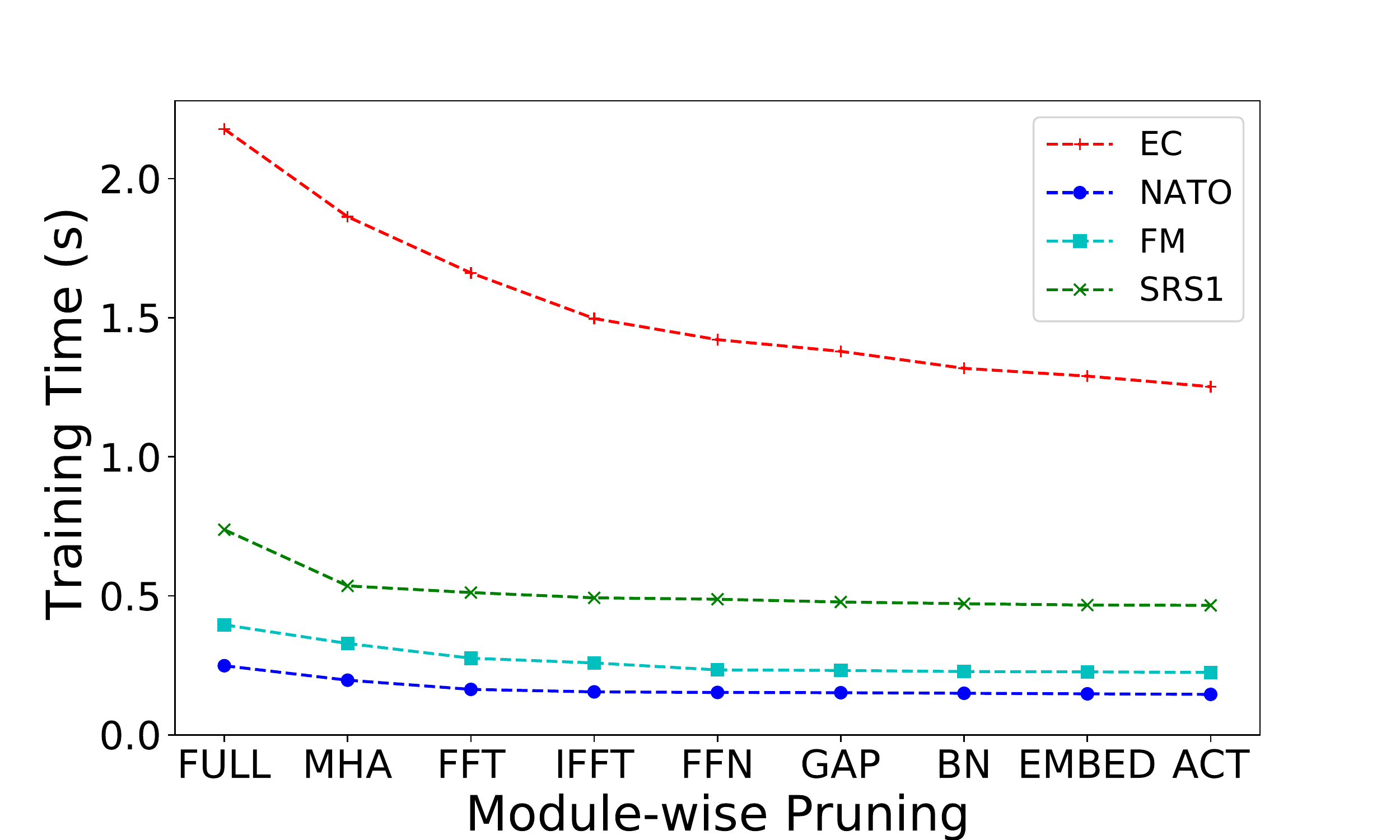}}
\caption{Module-wise results for changes in terms of number of parameters and training time per epoch on four datasets: EC, NATO, FM, SRS1.}
\label{change}
\end{figure}

\begin{figure}[htbp]
\centering
 \subfloat[]{\includegraphics[width=0.33\textwidth,]{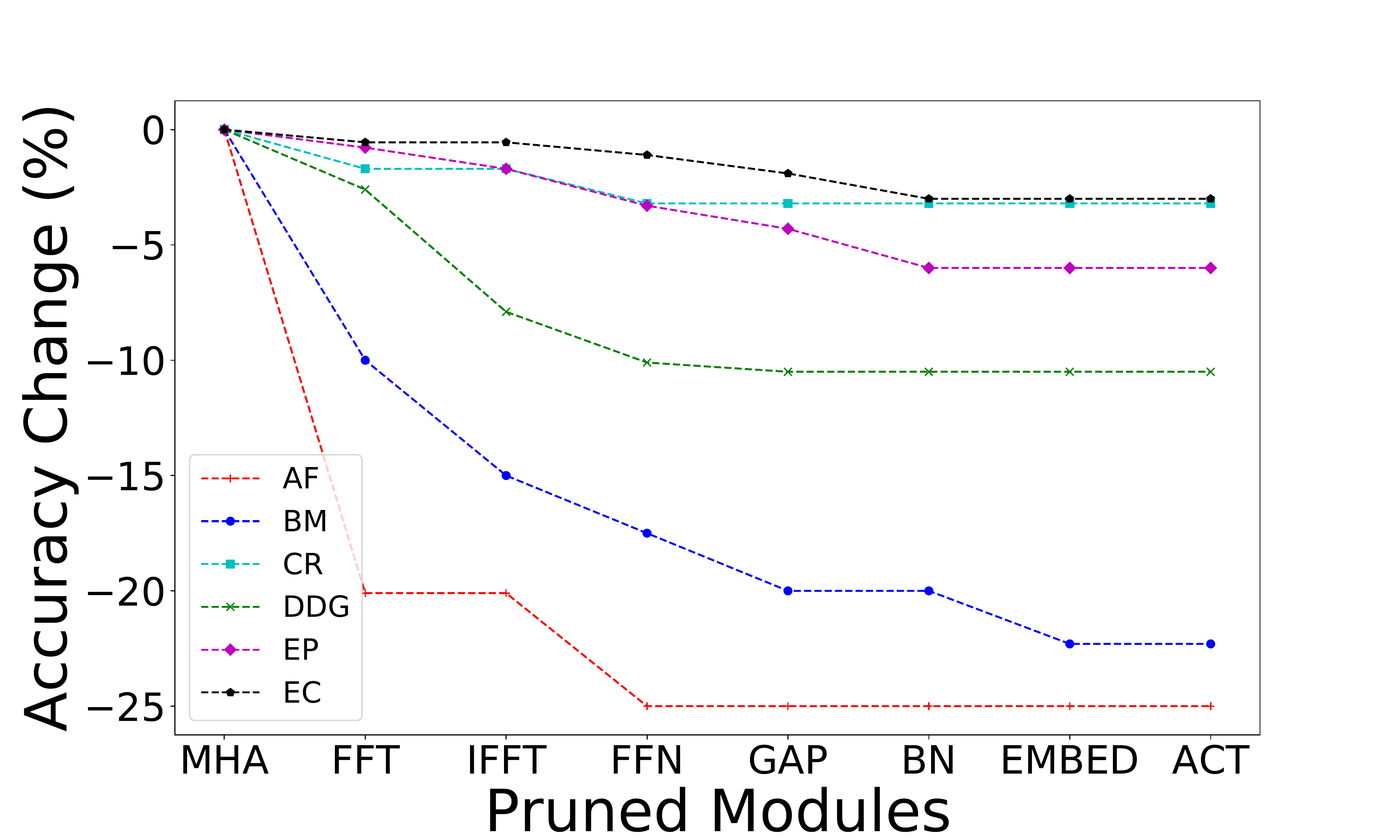}}
    \hfill
    \subfloat[]{\includegraphics[width=0.33\textwidth, ]{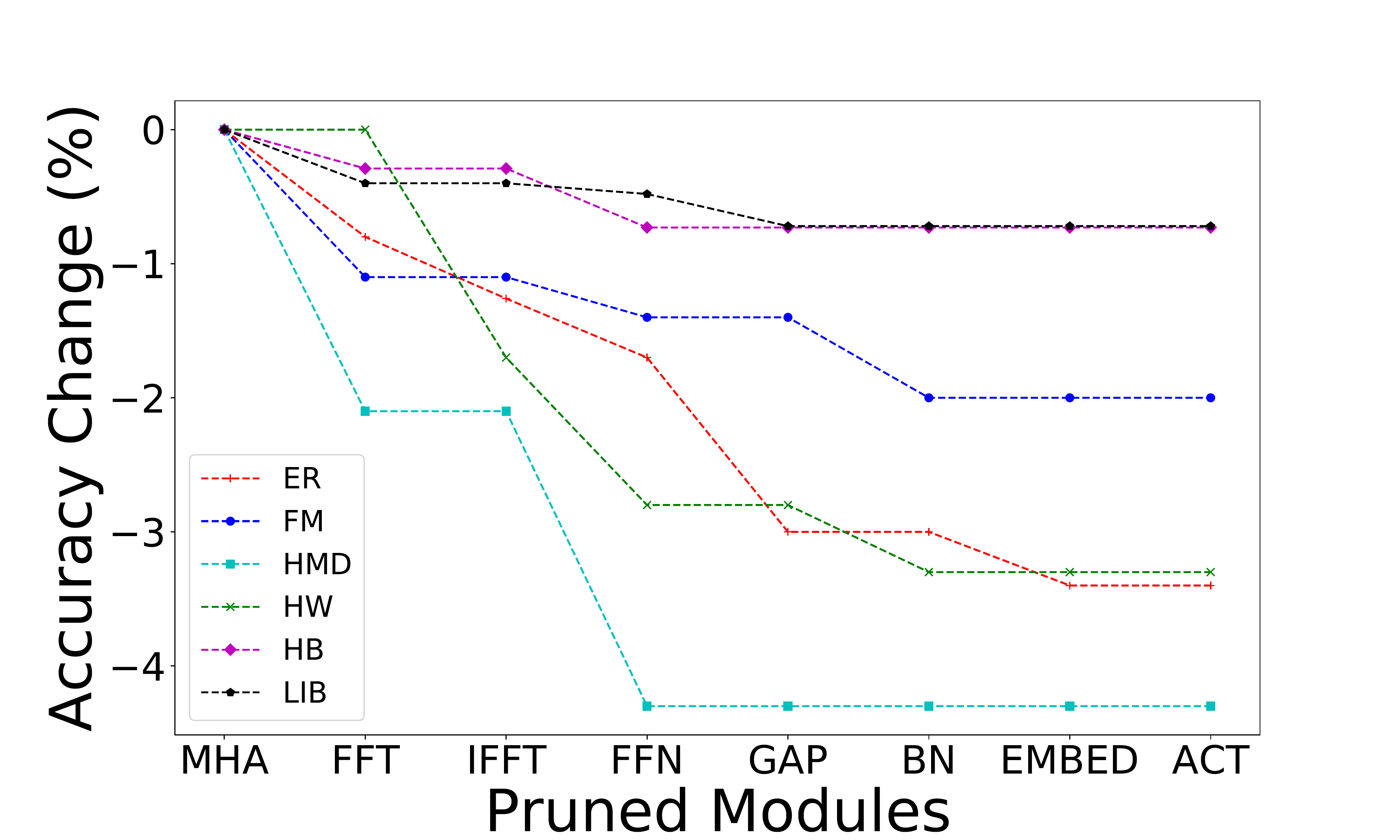}}
    \hfill
    \subfloat[]{\includegraphics[width=0.33\textwidth, ]{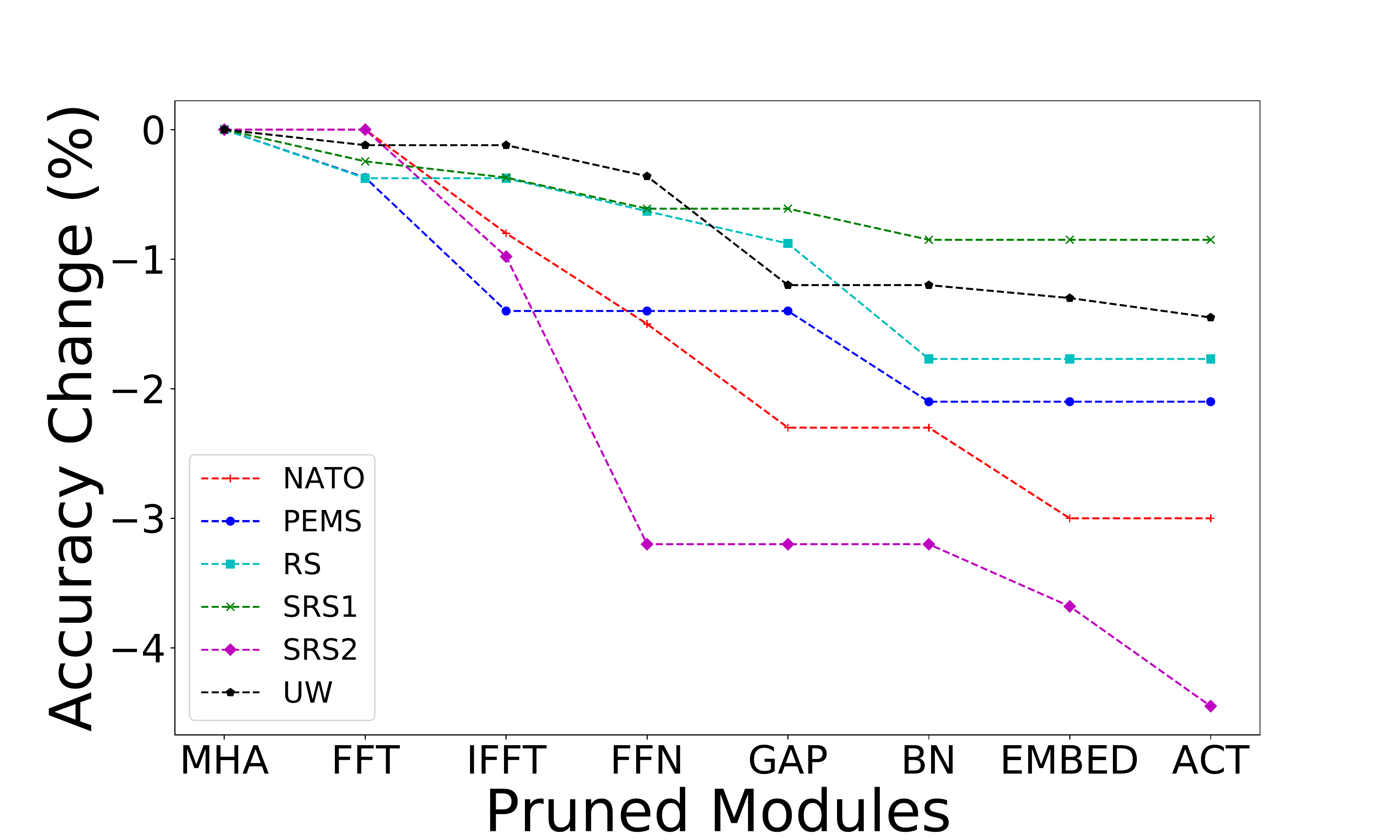}}
\caption{Change in accuracy (\%) from module-by-module pruning across all datasets. The order of datasets shown from (a) to (c) correspond to Table~\ref{ablation}.}
\label{pd}
\end{figure}

Next, we explore the relationship between efficiency (defined the same as Section~\ref{as}) and effectiveness (predictive performance). Based on the contribution of each module on the performance loss shown in Figure~\ref{acc_lossa}, we perform module-by-module pruning by following the order of modules from the most significant contributor to the least significant contributor (MHA, FFT, IFFT, FFN, GAP, BN, EMBED, ACT) to accuracy. We evaluate such pruning effect in two aspects: $(1)$
effectiveness: testing accuracy; $(2)$ efficiency: average training time per epoch in seconds and the number of learnable parameters. Due to limited space, we only show some datasets' testing accuracy in Table~\ref{prune1} and their efficiency results in Figure~\ref{change}. We observe that after removing the entire MHA module, the number of learnable parameters shrinks drastically, so as the accuracy (Table~\ref{prune1} Column 4). The representation capability of the pruned network, which has fewer parameters, is damaged since the amount of parameters is a key aspect to the network representation. Furthermore, the pace of accuracy loss and parameter reduction removal of subsequent modules slows down as FFT/IFFT has no learnable parameters. For the remaining modules, the number of parameters they carry is much fewer than the MHA module. Based on Figure~\ref{acc_lossa}, their effects on the predictive performance are moderate. Hence, the curves in Figure~\ref{change} are relatively flat following MHA. We further investigate the extent of change in accuracy of module-wise pruning on all datasets, as shown in Figure~\ref{pd}. We notice that the performance variation in different datasets vary widely. For datasets such as AF, BM, and DDG, the model pruning has a great impact on their performance. This may be due to very limit amount of training samples. Conversely, for datasets like HB, LIB, and SRS1, the model pruning brings little effect after removing the MHA module (within 1\%).

\begin{table}[H]
\centering
\caption{Module-wise pruning results of datasets EC, NATO, FM and SRS1. The results from Column $3$ (MHA) to Column $10$ (AF) with regard to accuracy represent that the module in that column is removed from the model architecture. Experiments are conducted $5$ times with different random seeds. The accuracy results are shown in the format of mean and standard deviation. Bold represents that the module brings about much accuracy loss compared to the unpruned model. Following MHA, the accuracy decreasing trend remains stable.}
\begin{tabular}{c| c c c c c c c c c c}
\Xhline{1.2pt}
\textbf{Dataset}  & \textbf{Acc.} & \textbf{Unpruned} & \textbf{MHA} & \textbf{FFT} &  \textbf{IFFT} & \textbf{FFN} &  \textbf{GAP} & \textbf{BN} & \textbf{IE} & \textbf{AF}  \\ 
\hline
\multirow{2}{*}{EC} & Mean &  0.456 & \textbf{0.365}  & 0.363  & 0.363  & 0.361  & 0.358  & 0.354  & 0.354  & 0.354 \\ 
&  Std. &   0.003 &  0.004 &  0.004 &  0.002 &  0.004 &  0.003 & 0.003 &  0.003 &  0.003 \\ 
\hline
\multirow{2}{*}{NATO}  & Mean & 0.844& \textbf{0.750}  & 0.750  & 0.744 & 0.739  & 0.733  & 0.733 & 0.728  & 0.728 \\ 
&  Std. &  0.005 &  0.006 &  0.003 &  0.004 &  0.006 & 0.005 & 0.006 &  0.004 &  0.005 \\ 
\hline
\multirow{2}{*}{FM} & Mean & 0.640 & \textbf{0.500} & 0.495  & 0.495  & 0.493  & 0.493  & 0.490  & 0.490  & 0.490  \\ 
&  Std. &  0.009 &  0.010 &  0.011 &  0.010 &  0.008 &  0.009 &  0.011 &  0.010 & 0.011 \\ 
\hline
\multirow{2}{*}{SRS1} & Mean & 0.915 & \textbf{0.819} & 0.817 & 0.816 & 0.814 & 0.814 & 0.812 & 0.812 & 0.812 \\ 
&  Std. &  0.005 & 0.005 &  0.003 &  0.003 & 0.0.004 &  0.006 &  0.003 &  0.004 &  0.005 \\ 

\Xhline{1.2pt}
\end{tabular}
\label{prune1}
\end{table}

\vspace{-1cm}

\vspace{-0.5cm}
Overall, based on the above module-by-module pruning scheme, we observe that as the effectiveness (predictive performance) of the network increases, the corresponding efficiency (training speed and model size) generally decreases. The evident cost--benefit trade-off between efficiency and effectiveness provides a key question to researchers on how to find efficient model settings while maintaining the “equilibrium" between these two aspects. This problem will be discussed in Section~\ref{par}.

\subsection{EFFICIENCY VS. COMPLEXITY}
\label{evc}
\begin{figure}[H]
    \centering
    \subfloat[]{\includegraphics[width=0.33\textwidth,]{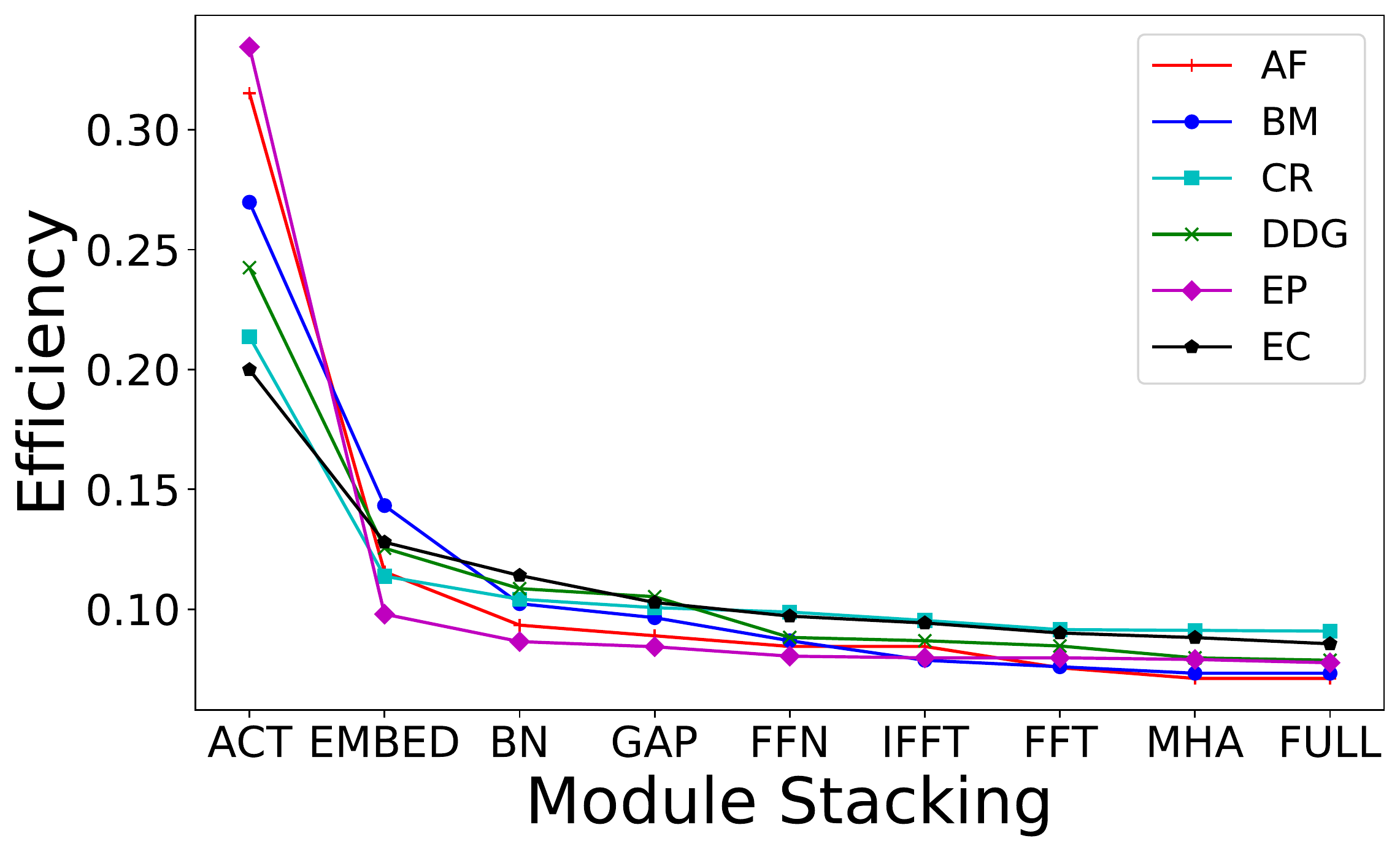}}
    \hfill
    \subfloat[]{\includegraphics[width=0.33\textwidth, ]{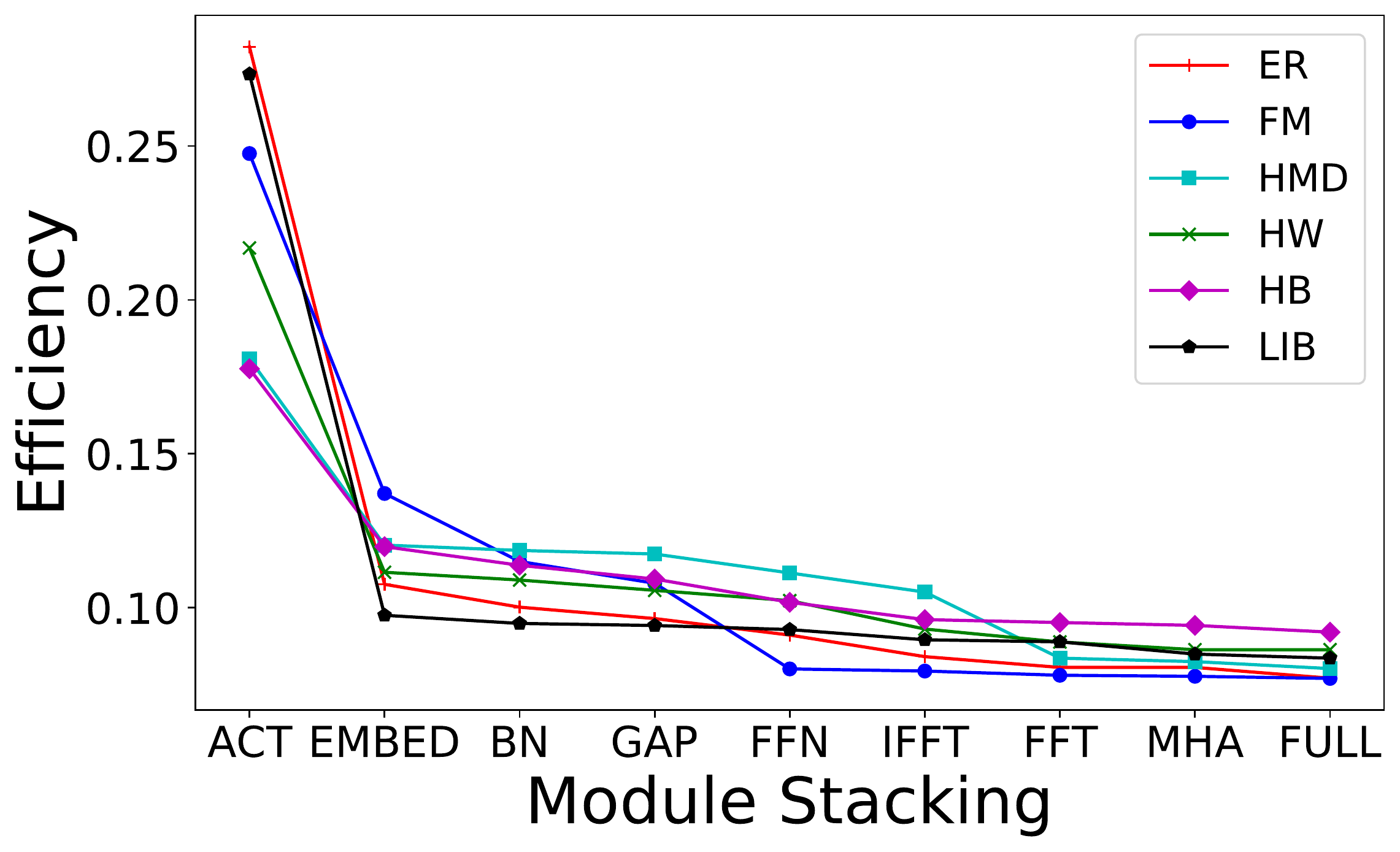}}
    \hfill
    \subfloat[]{\includegraphics[width=0.33\textwidth, ]{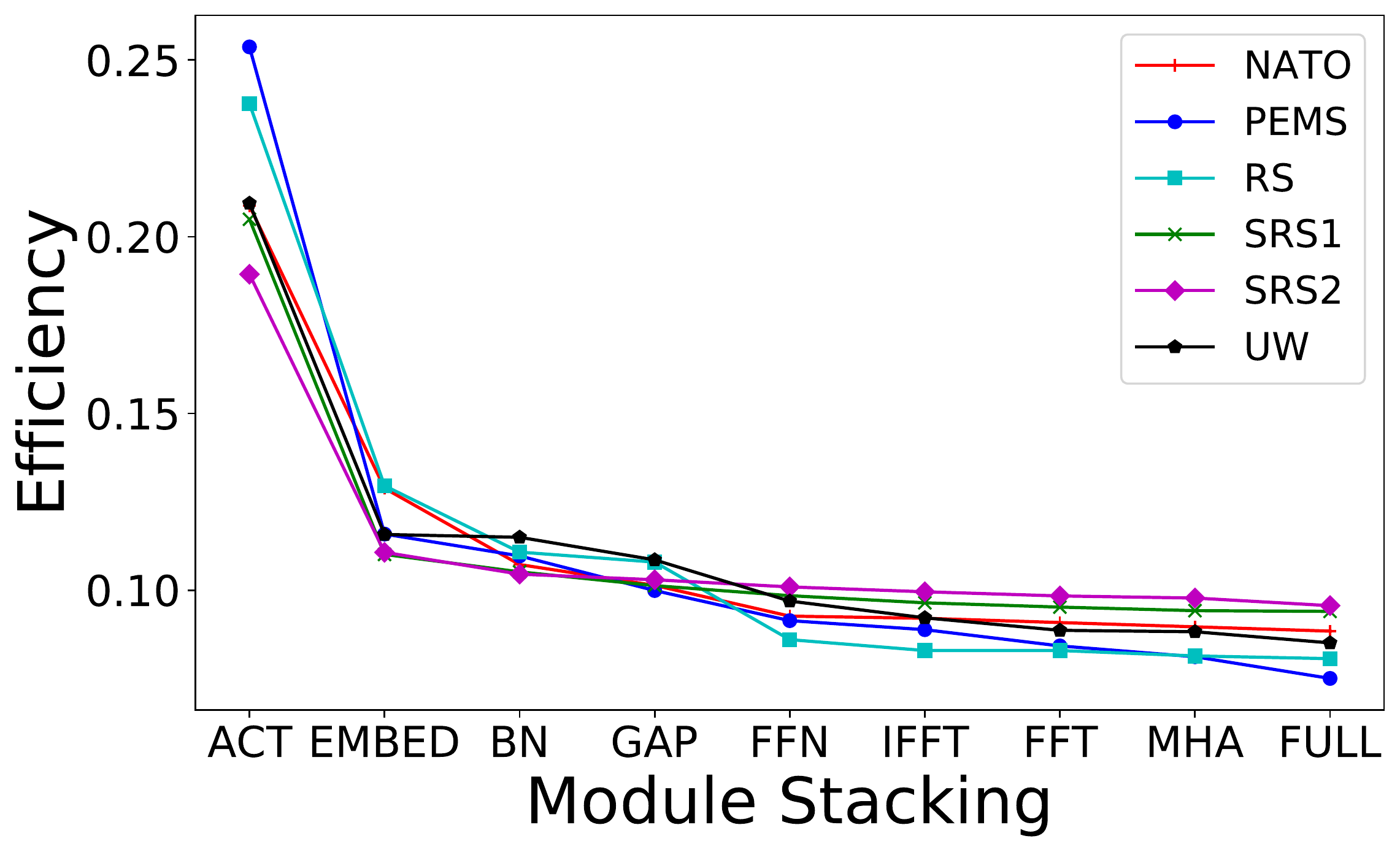}}
  \caption{Trade-off between network efficiency and complexity across all datasets. Due to the notable differences of dataset sizes, the computation of efficiency is normalized for each dataset. The order of datasets shown from (a) to (c) correspond to Table~\ref{ablation}.}
  \label{efftho}
\end{figure}

Here, we explore the relationship between network efficiency and complexity. In general, the more complex a model is, the less efficient it is. The network's efficiency is defined in the same way as previous sections, in terms of the training time and the number of parameters. Meanwhile, we define the complexity of the model as the stacking of modules. Contrary to model pruning, we stack each module based on their influence on the predictive performance, from the least significant contributor to the most significant contributor (ACT, EMBED, BN, GAP, FFN, IFFT, FFT, MHA) to accuracy. Our empirical results in Figure~\ref{efftho} shed light on the trade-off between model efficiency and complexity. As can be seen in Figure~\ref{efftho}, as more modules are stacked over the network, the corresponding computational efficiency decreases. All datasets illustrate similar trends.

\subsection{PARETO ANALYSIS FOR TRADE-OFF EXPLORATION BETWEEN EFFICIENCY AND PERFORMANCE/EFFECTIVENESS}
\label{par}
\begin{figure}[htbp]
    \centering
    \subfloat[\textit{FingerMovements}.]{\includegraphics[width=0.48\textwidth,]{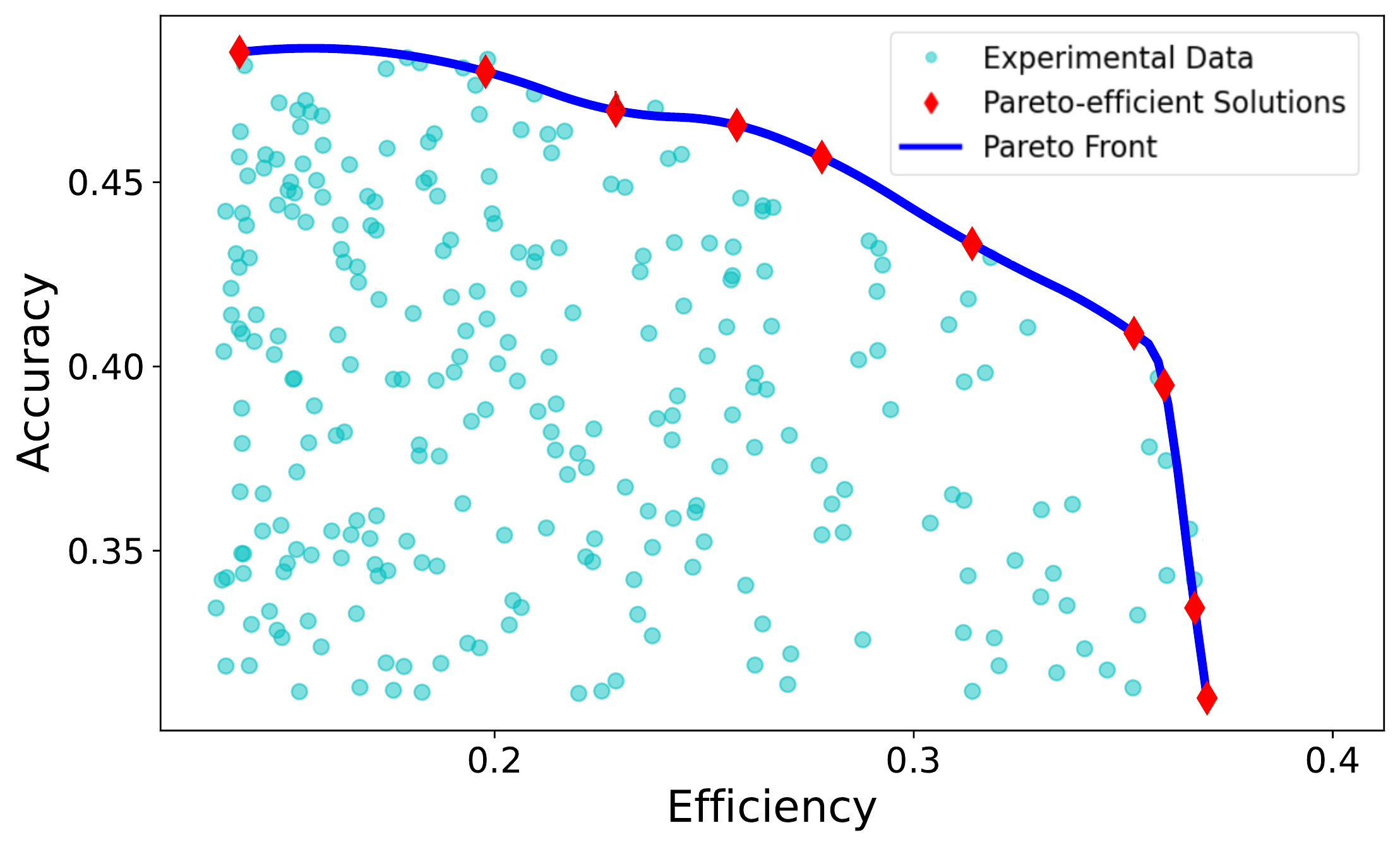}}
    \hfill
    \subfloat[\textit{Heartbeat}.]{\includegraphics[width=0.48\textwidth, ]{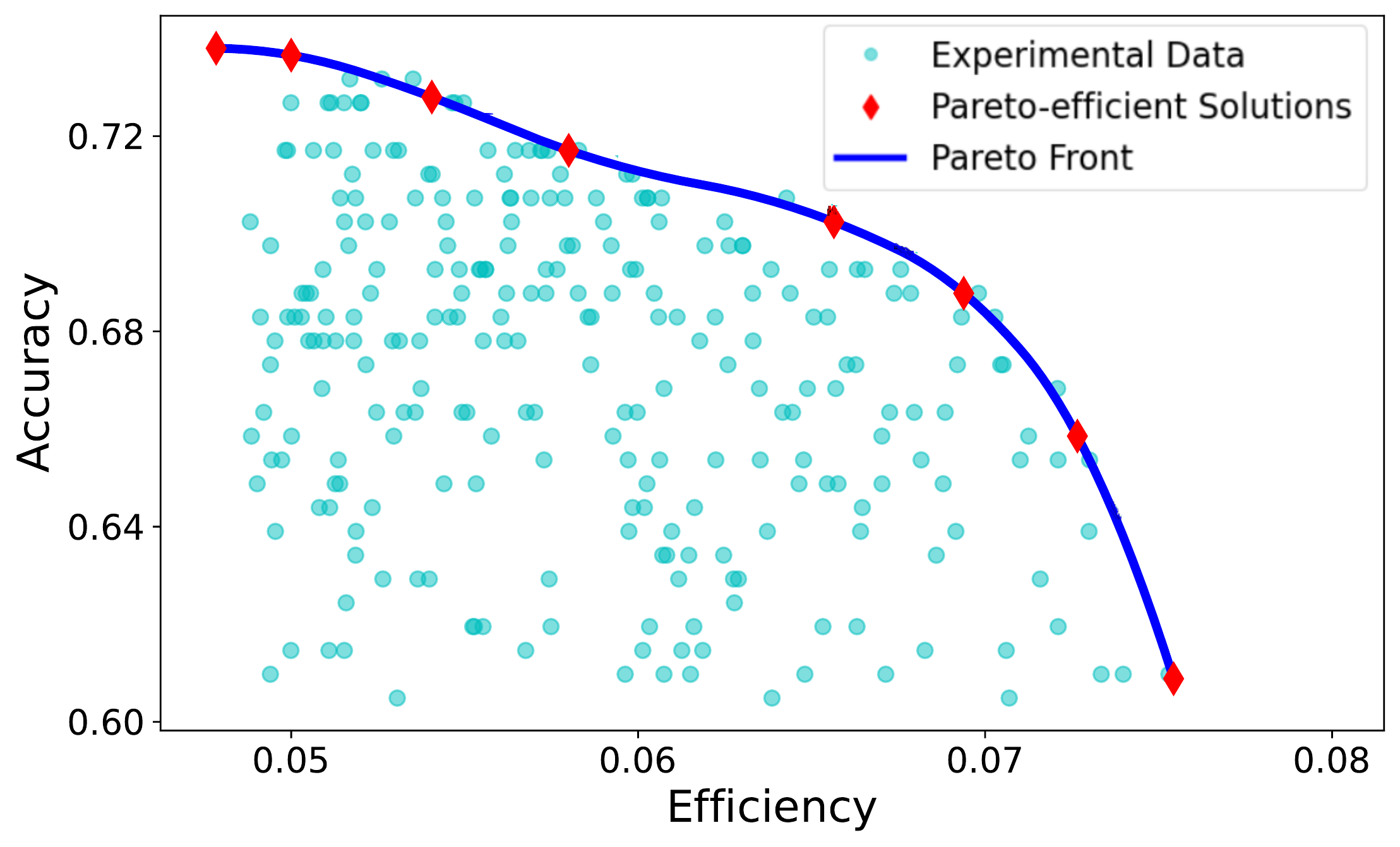}}
  \caption{Pareto efficiency visualization of the \textit{FingerMovements} and \textit{Heartbeat} datasets. The scattered cyan points, the marked red points, and the blue curve represent randomly sampled experimental data, Pareto-efficient solutions, and Pareto efficient frontiers.}
  \label{pareto}
\end{figure}

We define the model efficiency in terms of the reciprocal of the product between training time per epoch and the number of parameters. Thus, the higher the reciprocal, the higher the efficiency. To explore the relationship between model efficiency and performance, we employ Pareto analysis~\cite{censor1977pareto}. Pareto efficiency represents a state for which improving the performance as measured by one criterion would worsen the performance as measured by another criterion. 
We choose the \textit{FingerMovements} and \textit{Heartbeat} datasets to obtain the Pareto frontiers, where the set of points on the front correspond to Pareto-efficient solutions. We have two objectives: $(1)$ maximize the efficiency; $(2)$ maximize the accuracy. Figure~\ref{pareto} shows the result of Pareto fronts for both datasets in blue, where the red points are Pareto-efficient solutions. The scattered cyan points are randomly sampled experimental data from all different configurations. The Pareto analysis provides us with a principled approach for choosing efficient network settings, while exploring the trade-off between efficiency and performance. Specifically, we can identify the extent of computational resources that is required in order for a model to achieve a certain performance. Conversely, we can identify how well a model can perform, given a certain amount of resources.

\section{DISCUSSION}
\label{con}
In this work, we propose a methodology to investigate the relationship between model efficiency and effectiveness, as well as its complexity. The method is performed on a mixing network based on Transformer and Fourier transform for MTS classification. Extensive experiments are conducted on 18 MTS datasets, including ablation studies on different modules of the network, module-by-module pruning evaluated in terms of the predictive performance, training speed, and the number of learnable parameters. The network achieves competitive performance compared to current best-performing methods. Ablation studies indicate that self-attention and Fourier transform are the largest contributors that influence the model performance across all datasets. Furthermore, through sequential pruning of each module, we observed the efficiency--effectiveness and the efficiency--complexity trade-offs of the network. Through Pareto analysis, we show how to choose efficient settings of the network, while investigating the performance--efficiency trade-off through visualization of the Pareto fronts. We note that for far more complex models applied to large-scale data, due to finite computational resources, it is not practical to consider all possible configurations of the model and perform experiments. In these cases, given a reasonable number of experiments, techniques like regression can be used to generate massive random model settings and corresponding model performance. Pareto analysis can then be performed to evaluate the efficiency-performance trade-off, to guide researchers to adjust the model settings to improve the efficiency and effectiveness accordingly.

%
%
%
%

{}

\end{document}